\documentclass[11pt,a4paper]{article}
\usepackage[hyperref]{emnlp-ijcnlp-2019}
\usepackage{cmap}
\usepackage[LAE,LFE,T2A,T1]{fontenc}
\usepackage{times}
\usepackage{latexsym}
\usepackage{color}
\usepackage{multirow}
\usepackage{listings}
\usepackage{soul}
\usepackage{url}
\usepackage{fixltx2e}
\usepackage{graphicx}
\usepackage{subfigure}
\usepackage[utf8]{inputenc}
\usepackage[main=english,russian,arabic]{babel}
\usepackage{CJKutf8}
\usepackage{verbatim}
\usepackage{placeins}

\newcommand\blfootnote[1]{%
  \begingroup
  \renewcommand\thefootnote{}\footnote{#1}%
  \addtocounter{footnote}{-1}%
  \endgroup
}

\definecolor{codegreen}{rgb}{0,0.6,0}
\definecolor{codegray}{rgb}{0.5,0.5,0.5}
\definecolor{codepurple}{rgb}{0.58,0,0.82}
\definecolor{paperred}{rgb}{.84,0,0,}
\definecolor{greybackground}{rgb}{.98,0.98,0.98}

\lstdefinestyle{mystyle}{
    backgroundcolor=\color{greybackground},   
    commentstyle=\color{codegreen},
    keywordstyle=\color{purple},
    numberstyle=\tiny\color{codegray},
    stringstyle=\color{paperred},
    breakatwhitespace=false,         
    breaklines=true,                 
    captionpos=b,                    
    keepspaces=true,                 
    numbersep=5pt,                  
    showspaces=false,                
    showstringspaces=false,
    showtabs=false,                  
    tabsize=2,frame=single,
    basicstyle=\scriptsize\ttfamily,
    belowskip=-1.5em,
}
\aclfinalcopy 

\title{\textbf{Multilingual Universal Sentence Encoder\\for Semantic Retrieval}}

\author{\textbf{Yinfei Yang}\textsuperscript{$a\dagger$}, \textbf{Daniel Cer}\textsuperscript{$a\dagger$}, \textbf{Amin Ahmad}\textsuperscript{$a$}, \textbf{Mandy Guo}\textsuperscript{$a$}, \\
\textbf{Jax Law}\textsuperscript{$a$}, \textbf{Noah Constant}\textsuperscript{$a$}, \textbf{Gustavo Hernandez Abrego}\textsuperscript{$a$}, \textbf{Steve Yuan}\textsuperscript{$b$}, \textbf{Chris Tar}\textsuperscript{$a$}, \\
\textbf{Yun-Hsuan Sung}\textsuperscript{$a$}, \textbf{Brian Strope}\textsuperscript{$a$}, \textbf{Ray Kurzweil}\textsuperscript{$a$}  \AND
  {\rm\textsuperscript{$a$}Google AI}\\Mountain View, CA \And
  {\rm\textsuperscript{$c$}Google}\\Cambridge, MA
}

\date{}

\begin{document}

\maketitle

\begin{abstract}
We introduce two pre-trained retrieval focused multilingual sentence encoding models, respectively based on the Transformer and CNN model architectures.
The models embed text from \emph{16 languages} into a single semantic space using a multi-task trained dual-encoder that learns tied representations using translation based bridge tasks~\cite{mutty2018}.
The models provide performance that is competitive with the state-of-the-art on: semantic retrieval~(SR), translation pair bitext retrieval~(BR) and retrieval question answering~(ReQA).
On English transfer learning tasks, our sentence-level embeddings approach, and in some cases exceed, the performance of monolingual, English only, sentence embedding models. Our models are made available for download on TensorFlow Hub.

\blfootnote{$\dagger$ Corresponding authors:\\
{\tt \hphantom{\{yin}\{yinfeiy, cer\}@google.com}}

\end{abstract}

\thispagestyle{empty}
\section{Introduction}

We introduce three new members in the \emph{universal sentence encoder} (USE) \cite{use2018} family of sentence embedding models. Two multilingual models, one based on CNN~\cite{kim-2014-convolutional} and the other based on the Transformer architecture~\cite{vaswani2017}, target performance on tasks requiring models to capture multilingual semantic similarity. The third member introduced is an alternative interface to our multilingual Transformer model for use in retrieval question answering~(ReQA). The \emph{16 languages} supported by our multilingual models are given in Table \ref{tab:langs}.\footnote{Due to character set differences, we treat Simplified Chinese, zh, and Traditional Chinese, zh-tw, prominently used in Taiwan, as two languages within our model.}

\begin{table}[!htb]
    \centering
    \footnotesize
    \begin{tabular}{|l|l|l|l|}
        \hline
        \textbf{Languages} & \textbf{Family} \\
        \hline
        Arabic (ar)  & \textbf{Semitic} \\ 
        \hline
        Chinese (PRC) (zh) &  \textbf{Sino-Tibetan} \\
        Chinese (Taiwan) (zh-tw) & \\
        \hline
        Dutch(nl) English(en) 
        & \textbf{Germanic} \\
        German (de) & \\
        \hline
        French (fr) Italian (it)  
        & \textbf{Latin} \\
        Portuguese (pt) Spanish (es) & \\
        \hline
        Japanese (ja) & \textbf{Japonic} \\
        \hline
        Korean (ko) & \textbf{Koreanic} \\
        \hline
        Russian (ru) Polish (pl) & \textbf{Slavic} \\
        \hline
        Thai (th) & \textbf{Kra–Dai} \\
        \hline
        Turkish (tr) & \textbf{Turkic} \\
        \hline
    \end{tabular}
    \caption{Supported languages (ISO 639-1).}
    \label{tab:langs}
\end{table}

\section{Model Toolkit}

Models are implemented in TensorFlow \cite{tf2016} and made publicly available on TensorFlow Hub.\footnote{ \url{https://www.tensorflow.org/hub/}, Apache 2.0 license, with models available as saved TF graphs.} Listing \ref{uesnippet} illustrates the generation of sentence embeddings using one of our multilingual models. Listing \ref{qasnippet} demonstrates using the question answering interface. Responses are encoded with additional context information such that the resulting embeddings have a high dot product similarity score with the questions they answer. This allows for retrieval of indexed candidates using efficient nearest neighbor search.\footnote{Popular efficient search tools include FAISS \url{https://github.com/facebookresearch/faiss}, Annoy \url{https://github.com/spotify/annoy}, or FLANN \url{https://www.cs.ubc.ca/research/flann}.}

\lstset{framesep=5pt}
\begin{lstlisting}[language=Python,label=uesnippet,float=htb, escapeinside={*}{*},caption=Encoding for STS/Bitext retrieval.]
import tensorflow_hub as hub

module = hub.Module("https://tfhub.dev/google/"
    "universal-sentence-encoder-multilingual/1")

multilingual_embeddings = module([ 
 "Hola Mundo!", "Bonjour le monde!", "Ciao mondo!"
 "Hello World!", "Hallo Welt!", "Hallo Wereld!",
 "*{\color{red}\begin{CJK}{UTF8}{min}你好世界!\end{CJK}}*", "*\foreignlanguage{russian}{\color{red}Привет, мир!}*", "*\foreignlanguage{arabic}{\color{red}\AR{مرحبا بالعالم!}}*"])

\end{lstlisting}

\lstset{framesep=5pt}

\begin{lstlisting}[language=Python,label=qasnippet,float=htb,caption=Encoding for QA retrieval.]
module = hub.Module("https://tfhub.dev/google/"
    "universal-sentence-encoder-multilingual-qa/1")

query_embeddings = module(
    dict(text=["What is your age?"]),
    signature="question_encoder", as_dict=True)
  
candidate_embeddings = module(
    dict(text=["I am 20 years old."],
         context=["I will be 21 next year."]),
    signature="response_encoder", as_dict=True)

\end{lstlisting}

\section{Encoder Architecture}

\subsection{Multi-task Dual Encoder Training}
Similar to \citet{use2018} and \citet{mutty2018}, we target broad coverage using a multi-task dual-encoder training framework, with a single shared encoder supporting multiple downstream tasks.
The training tasks include: a multi-feature  question-answer prediction task,\footnote{Question-answer prediction is similar to conversational-response prediction~\cite{yang2018}. We treat the question as the conversational input and the answer as the response. For improved answer selection, we provide a bag-of-words (BoW) context feature as an additional input to the answer encoder. The context could be the surrounding text or longer version of answer that provides more information. The context feature is encoded using a separate DAN encoder.} a translation ranking task, and a natural language inference (NLI) task.
Additional task specific hidden layers for the question-answering and NLI tasks are added after the shared encoder to provide representational specialization for each type of task.

\subsection{SentencePiece}
SentencePiece tokenization~\cite{spm} is used for all of the 16 languages supported by our models.
A single 128k SentencePiece vocabulary is trained from 8 million sentences sampled from our training corpus and balanced across the 16 languages.
For validation, the trained vocab is used to process a separate development set, also sampled from the sentence encoding model training corpus.
We find the character coverage is higher than 99\% for all languages, which means less than 1\% output tokens are out of vocabulary.
Each token in the vocab is mapped to a fixed length embedding vector.\footnote{Out-of-vocabulary characters map to an \texttt{<UNK>} token.}

\subsection{Shared Encoder}

Two distinct architectures for the sentence encoding models are provided: (i) transformer \cite{vaswani2017}, targeted at higher accuracy at the cost of resource consumption; (ii) convolutional neural network (CNN)~\cite{kim-2014-convolutional}, designed for efficient inference but obtaining reduced accuracy.

\paragraph{Transformer}

The transformer encoding model embeds sentences using the \emph{encoder} component of the transformer architecture~\cite{vaswani2017}. Bi-directional self-attention is used to compute context-aware representations of tokens in a sentence, taking into account both the ordering and the identity of the tokens. The context-aware token representations are then averaged together to obtain a sentence-level embedding.

\paragraph{CNN}

The CNN sentence encoding model feeds the input token sequence embeddings into a convolutional neural network~\cite{kim-2014-convolutional}. Similar to the transformer encoder, average pooling is used to turn the token-level embeddings into a fixed-length representation. Sentence embeddings are then obtain by passing the averaged representation through additional feedforward layers.

\section{Training and Configuration}

\subsection{Training Corpus}

Training data consists of mined question-answer pairs,\footnote{QA pairs are mined from online forums and QA websites, including Reddit, StackOverflow, and YahooAnswers.} mined translation pairs,\footnote{The translation pairs are mined using a system similar to the approach described in \citet{jakob2010}.} and the Stanford Natural Language Inference (SNLI) corpus~\cite{bowman-etal-2015-large}.\footnote{MultiNLI~\cite{williams-etal-2018-broad}, a more extensive corpus, contains examples from multiple sources but with different licences. Employing SNLI avoids navigating the licensing complexity of using MultiNLI to training public models.}
SNLI only contains English data. The number of mined questions-answer pairs also varies across languages with a bias toward a handful of top tier languages.
To balance training across languages, we use Google's translation system to translate SNLI to the other 15 languages.
We also translate a portion of question-answer pairs to ensure each language has a minimum of 60M training pairs.
For each of our datasets, we use 90\% of the data for training, and the remaining 10\% for development/validation.

\subsection{Model Configuration}
Input sentences are truncated to 256 tokens for the CNN model and 100 tokens for the transformer. The CNN encoder uses 2 CNN layers with filter width of [1, 2, 3, 5] and a filter size of 256.
The Transformer encoder employs 6 transformer layers, with 8 attentions heads, hidden size 512, and filter size 2048.
Model hyperparameters are tuned on development data sampled from the same sources as the training data. We export sentence encoding modules for our two encoder architectures: \textbf{USE\textsubscript{Trans}} and \textbf{USE\textsubscript{CNN}}. We also export a larger graph for QA tasks from our Transformer based model that includes QA specific layers and support providing context information from the larger document as \textbf{USE\textsubscript{QA Trans+Cxt}}.\footnote{While USE\textsubscript{QA Trans+Cxt} uses the same underlying shared encoder as USE\textsubscript{Trans} but with additional task specific layers, we anticipate that the models could diverge in the future.}

\begin{table}
    \small
    \centering
    \begin{tabular}{l | r  r r}
         \textbf{Model} &  \textbf{Quora} & \textbf{AskUbuntu} & \textbf{Average}\\ \hline
         \citet{Gillick2018EndtoEndRI} & 87.5 & 37.3 & 62.4  \\
         USE\textsubscript{CNN}      & 89.2 & 39.9 & 64.6  \\
         USE\textsubscript{Trans}  & 89.1 & 42.3 & 65.7  \\
    \end{tabular}
    \caption{MAP@100 on SR (English). Models are compared with the best models from \citet{Gillick2018EndtoEndRI} that do not benefit from in-domain training data.}
    \label{tab:re_sts}
\end{table}

\section{Experiments on Retrieval Tasks}

In this section we evaluate our multilingual encoding models on semantic retrieval, bitext and retrieval question answer tasks.

\subsection{Semantic Retrieval (SR)}

Following \citet{Gillick2018EndtoEndRI}, we construct semantic retrieval (SR) tasks from the Quora question-pairs~\cite{quora} and AskUbuntu~\cite{askubuntu} datasets. The SR task is to identify all sentences in the retrieval corpus that are semantically similar to a query sentence.\footnote{The task is related to paraphrase identification \cite{dolan-etal-2004-unsupervised} and Semantic Textual Similarity (STS) \cite{cer-etal-2017-semeval}, but with the identification of meaning similarity being assessed in the context of a retrieval task.}    

For each dataset, we first build a graph connecting each of the positive pairs, and then compute its transitive closure.
Each sentence then serves as a test query that should retrieve all of the other sentences it is connected to within the transitive closure. Mean average precision (MAP) is employed to evaluate the models.
More details on the constructed datasets can be found in \citet{Gillick2018EndtoEndRI}.
Both datasets are English only.

Table \ref{tab:re_sts} shows the MAP@100 on the Quora/AskUbuntu retrieval tasks.
We use \citet{Gillick2018EndtoEndRI} as the baseline model, which is trained using a similar dual encoder architecture.
The numbers listed here are from the models without in-domain training data~\footnote{The model for Quora is trained on Paralex (\url{http://knowitall.cs.washington.edu/paralex}) and AskUbuntu data. The model for AskUbuntu is trained on Paralex and Quora.}.

\begin{table}
    \small
    \centering
    \begin{tabular}{l| r r r r}
        \textbf{Model} & \textbf{en-es} & \textbf{en-fr} & \textbf{en-ru} & \textbf{en-zh} \\ \hline
         \citet{Yang2019ImprovingMS} & 89.0 & 86.1 & 89.2 & 87.9 \\
         USE\textsubscript{CNN}      & 85.8 & 82.7 & 87.4 & 79.5  \\
         USE\textsubscript{Trans}  & 86.1 & 83.3 & 88.9 & 78.8 \\
    \end{tabular}
    \caption{P@1 on UN Bitext retrieval task.}
    \label{tab:re_bitext}
\end{table}

\subsection{Bitext Retrieval (BR)}

Bitext retrieval performance is evaluated on the United Nation (UN) Parallel Corpus~\cite{uncorpus}, containing 86,000 bilingual document pairs matching English (en) documents with with their translations in five other languages: French (fr), Spanish (es), Russian (ru), Arabic (ar) and Chinese (zh).
Document pairs are aligned at the sentence-level, which results in 11.3 million aligned sentence pairs for each language pair.

Table \ref{tab:re_bitext} shows precision@1~(P@1) for the proposed models as well as the current state-of-the-art results from~\citet{Yang2019ImprovingMS},
which uses a dual-encoder architecture trained on mined bilingual data.
USE\textsubscript{Trans} is generally better than USE\textsubscript{CNN}, performing lower than the SOTA but not by too much with the exception of en-zh.\footnote{Performance is degraded from~\citet{Yang2019ImprovingMS} due to using a single sentencepiece vocabulary to cover 16 languages. Languages like Chinese, Korean, Japanese have much more characters. To ensure the vocab coverage, sentencepiece tends to split the text of these languages into single characters, which increases the difficulty of the task.}

\subsection{Retrieval Question Answering (ReQA)}

\begin{table}
    \small
    \centering
    \begin{tabular}{l | r  r }
         \textbf{Model} &  \textbf{SQuAD Dev} & \textbf{SQuAD Train} \\ \hline
         \multicolumn{3}{c}{\emph{Paragraph Retrieval}}\\
         \hline
         USE\textsubscript{QA Trans+Cxt} & 63.5 & 53.3   \\
         BM25 (baseline)  & 61.6 & 52.4 \\
         \hline
         \multicolumn{3}{c}{\emph{Sentence Retrieval}}\\
         \hline
         USE\textsubscript{Trans} & 47.1 & 37.2 \\
         USE\textsubscript{QA Trans+Cxt}  & 53.2 & 43.3   \\
    \end{tabular}
    \caption{P@1 for SQuAD ReQA. Models are not trained on SQuAD.
    Dev and Train only refer to the respective sections of the SQuAD dataset.}
    \label{tab:squad_en}
\end{table}

\begin{table*}[htb!]
    \small
    \centering
    \begin{tabular}{l | c@{~~~} | c@{~~~} c@{~~~} c@{~~~} c@{~~~} c@{~~~} c@{~~~} c@{~~~} c@{~~~} c@{~~~} c@{~~~} c@{~~~} c@{~~~} c@{~~~} c@{~~~} }
         \textbf{Model} & \textbf{en} & \textbf{ar} & \textbf{de} & \textbf{es} & \textbf{fr} & \textbf{it} & \textbf{ja} & \textbf{ko} & \textbf{nl} & \textbf{pt}  & \textbf{pl} & \textbf{ru} & \textbf{th} & \textbf{tr} & \textbf{zh / zh-t} \\ \hline
    \multicolumn{16}{c}{\textit{Cross-lingual Semantic Retrieval (cl-SR)}} \\ \hline
    \rule{-2pt}{8pt}
         \textbf{Quora}   & & & & & & & & & & & & & & &\\
         ~~~~USE\textsubscript{CNN}         & 89.2 & 79.9 & 83.7 & 85.0 & 85.0 & 85.5 & 82.4 & 77.6 & 81.3 & 85.2 & 78.3 & 83.8 & 83.5 & 79.9 & 81.9 \\
         ~~~~USE\textsubscript{Trans} & 89.1 & 83.1 & 85.5 & 86.3 & 86.7 & 86.8 & 85.1 & 82.5 & 83.8 & 86.5 & 82.1 & 85.7 & 85.8 & 82.5 & 84.8 \\
         \textbf{AskUbuntu}   & & & & & & & & & & & & & & &\\
         ~~~~USE\textsubscript{CNN}         & 39.9 & 33.0 & 35.0 & 35.6 & 35.2 & 36.1 & 35.5 & 35.1 & 34.5 & 35.6 & 32.9 & 35.2 & 35.2 & 32.8 & 34.6 \\
         ~~~~USE\textsubscript{Trans} & 42.3 & 38.2 & 40.0 & 39.9 & 39.3 & 40.2 & 40.6 & 40.3 & 39.5 & 39.8 & 38.4 & 39.6 & 40.3 & 37.7 & 40.1\\
         \textbf{Average}   & & & & & & & & & & & & & & &\\
         ~~~~USE\textsubscript{CNN}         & 64.6 & 56.5 & 59.4 & 60.3 & 60.1 & 60.8 & 59.0 & 56.4 & 57.9 & 60.4 & 55.6 & 59.5 & 59.4 & 56.4 & 58.3 \\
         ~~~~USE\textsubscript{Trans} & 65.7 & 60.7 & 62.8 & 63.1 & 63.0 & 63.5 & 63.8 & 62.4 & 61.7 & 63.2 & 60.7 & 62.7 & 63.1 & 60.1 & 62.5 \\
    \hline
    \multicolumn{16}{c}{\textit{Cross-lingual Retrieval Question Answering (cl-ReQA)}} \\ \hline
    \rule{-2pt}{8pt}
         \textbf{SQuAD train} & & & & & & & & & & & & & & &\\
         ~~~~USE\textsubscript{QA Trans+Cxt}  & 43.3 & 33.2 & 35.2 & 37.2 & 37.0 & 37.0 & 32.9 & 31.1 & 36.6 & 37.7 & 34.5 & 33.2 & 36.9 & 32.3 & 32.7 \\
    \end{tabular}
    \caption{Cross-lingual performance on Quora/AskUbuntu cl-SR (MAP) and SQuAD cl-ReQA (P@1). Queries/questions are machine translated to the other languages, while retrieval candidates remain in English.}
    \label{tab:squad_xling}
\end{table*}

Similar to the data set construction used for the SR tasks, the SQuAD v1.0 dataset~\cite{squad} is transformed into a retrieval question answering (ReQA) task.\footnote{The retrieval question answering task was suggested by \citet{chen-etal-2017-reading} and then recently explored further by \citet{cakaloglu2018}. However, \citet{cakaloglu2018}'s use of sampling makes it difficult to directly compare with their results and we provide our own baseline base on BM25.} We first break all documents in the dataset into sentences using an off-the-shelf sentence splitter.
Each question of the (question, answer spans) tuples in the dataset is treated as a query.
The task is to retrieve the sentence designated by the tuple answer span.
Search is performed on a retrieval corpus consisting of all of the sentences within the corpus. We contrast sentence and paragraph-level retrieval using our models, with the later allowing for comparison against a BM25 baseline~\cite{bm25}.\footnote{BM25 is a strong baseline for text retrieval tasks. Paragraph-level experiments use the BM25 implementation: \url{https://github.com/nhirakawa/BM25}, with default parameters. We exclude sentence-level BM25, as BM25 generally performs poorly at this granularity.}

We evaluated ReQA using the SQuAD dev and train sets and without training on the SQuAD data.\footnote{
For sentences, the resulting retrieval task for development set consists of 11,425 questions and 10,248 candidates, and the retrieval task for train set is consists of 87,599 questions and 91,703 candidates. For paragraph retrieval, there are 2,067 retrieval candidates in the development set and 18,896 in the training set. To retrieve paragraphs with our model, we first run sentence retrieval and use the retrieved nearest sentence to select the enclosing paragraph.} The sentence and paragraph retrieval P@1 are shown in table \ref{tab:squad_en}. For sentence retrieval, we compare encodings produced using context from the text surrounding the retrieval candidate, USE\textsubscript{QA Trans+Cxt}, to sentence encodings produced without contextual cues, USE\textsubscript{Trans}. Paragraph retrieval contrasts USE\textsubscript{QA Trans+Cxt} with BM25.

\subsection{Cross-lingual Retrieval}

Our earlier experiments are extended to explore cross-lingual semantic retrieval (cl-SR) and cross-lingual retrieval question answering (cl-ReQA). SR queries and ReQA questions are machine translated into other languages, while keeping the retrieval candidates in English.\footnote{Poor translations are detected and rejected when the original English text and English back translation have a cosine similarity $< 0.5$ according our previously released English USE\textsubscript{Trans} model~\cite{use2018}.} Table \ref{tab:squad_xling} provides our cross-lingual retrieval results. On all the languages, USE\textsubscript{Trans} outperforms USE\textsubscript{CNN}. While cross-lingual performance lags the English only tasks, the performance is surprisingly close given the added difficulty of the cross-lingual setting.

\begin{table*}[hbt!]
\small
\centering
    \begin{tabular}{l @{~~~~~~}r @{~~~~~~}r r r r r@{~~} @{~~~~}r}
    \hline
    \multirow{2}{*}{\bf Model} &  \multirow{2}{*}{\bf MR} & \multirow{2}{*}{\bf CR} & \multirow{2}{*}{\bf SUBJ} & \multirow{2}{*}{\bf MPQA} & \multirow{2}{*}{\bf TREC} & \multirow{2}{*}{\bf SST} & {\bf STS Bench}\\
    & & & & & & & {\bf (dev / test)} \\
    \hline
    \rule{-2pt}{8pt}
    \textbf{USE mutlilingual models} & & & & & & \\
    ~~~~USE\textsubscript{CNN}                                      & 73.8 & 83.2 & 90.1 & 87.7 & 96.4 & 78.1 & 0.829 / 0.809 \\ 
    ~~~~USE\textsubscript{Transformer}                              & 78.1 & \textbf{87.0} & 92.1 & 89.9 & \textbf{96.6} & 80.9 & \textbf{0.837 / 0.825} \\ 
    \rule{-2pt}{8pt}
    \textbf{The state-of-the-art English embedding models} & & & & & &\\
    ~~~~InferSent~\cite{conneau2017}                                & 81.1 & 86.3 & 92.4 & \textbf{90.2} & 88.2 & 84.6 & 0.801 / 0.758 \\
    ~~~~Skip-Thought~~LN~\cite{ba2016}                              & 79.4 & 83.1 & 93.7 & 89.3 & -- & -- & -- \\
    ~~~~Quick-Thought~\cite{quickthought2018}                       & \textbf{82.4} & 86.0 & 94.8 & \textbf{90.2} & 92.4 & \textbf{87.6} & -- \\
    ~~~~USE\textsubscript{DAN}~~for English~\cite{use2018}          & 72.2 & 78.5 & 92.1 & 86.9 & 88.1 & 77.5 & 0.760 / 0.717 \\
    ~~~~USE\textsubscript{Transformer}~~for English~\cite{use2018}  & 82.2 & 84.2 & \textbf{95.5} & 88.1 & 93.2 & 83.7 & 0.802 / 0.766 \\
    \hline
    \end{tabular}
\caption{Performance on English transfer tasks from SentEval \citep{conneau2018senteval}.}
\label{tab:trans-model-performance}
\end{table*}

\begin{figure*}[ht!]
    \centering
    \subfigure[CPU Inference Time]{
        \includegraphics[width=4.9cm,trim=4 4 4 4,clip]{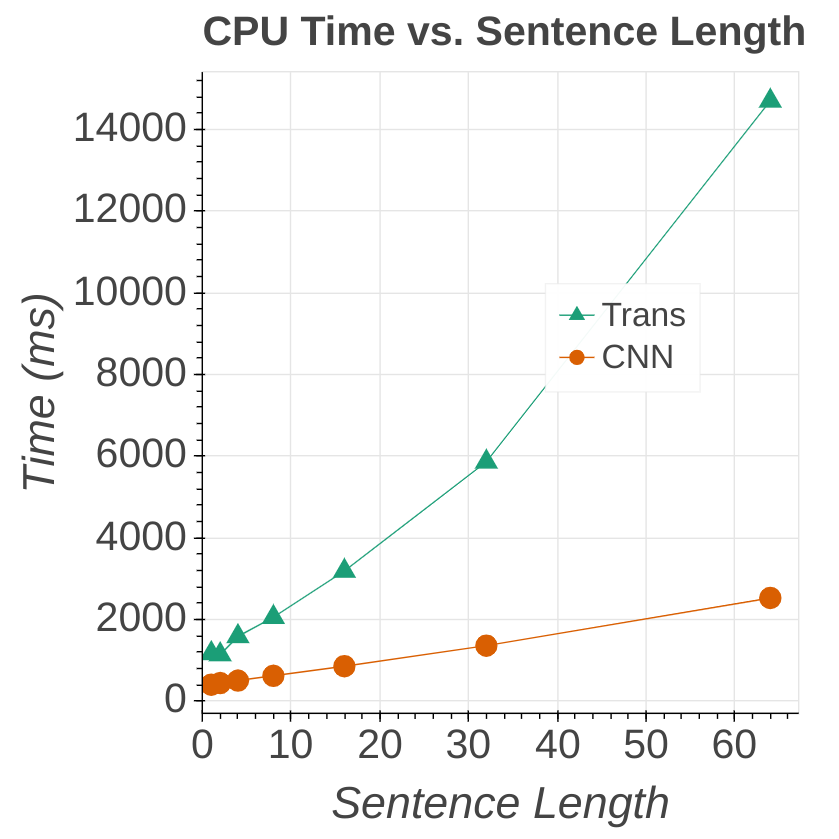}
        \label{fig:benchmarks:cputime}
    }
    \subfigure[GPU Inference Time]{
        \includegraphics[width=4.9cm,trim=4 4 4 4,clip]{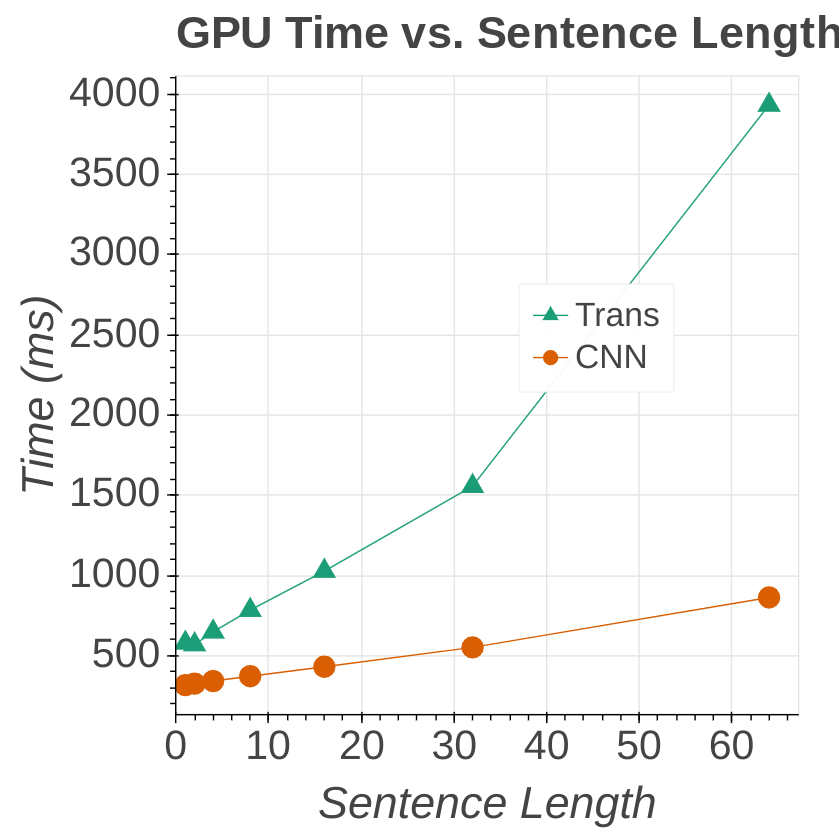}
        \label{fig:benchmarks:gputime}
    }
    \subfigure[Memory Footprint]{
        \includegraphics[width=4.9cm,trim=4 4 4 4,clip]{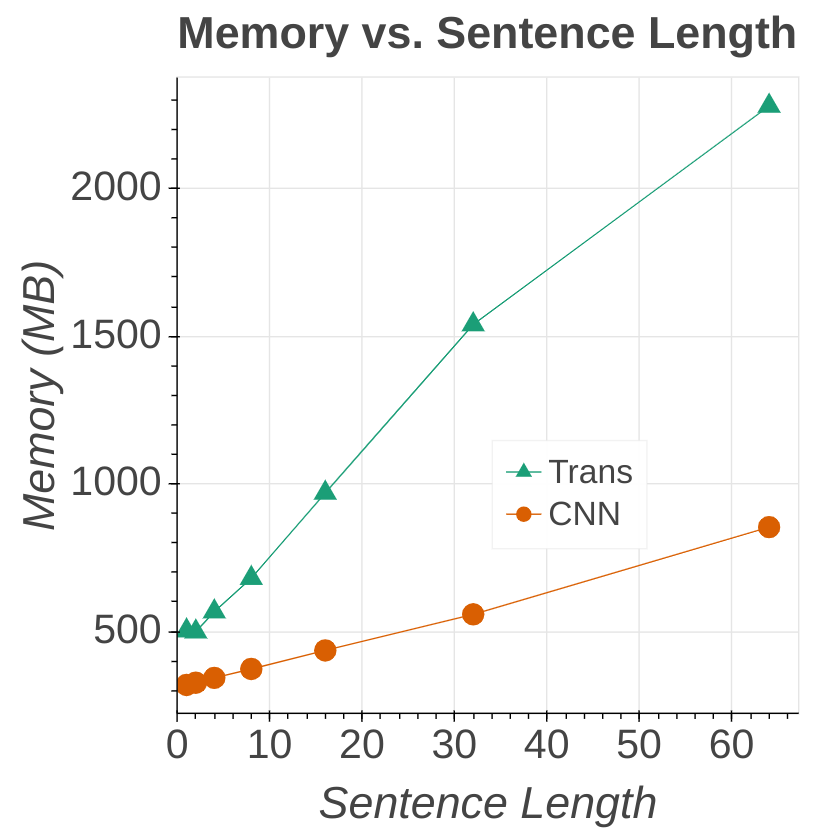}
        \label{fig:benchmarks:mem}
    }
    \caption{Resource usage for the multilingual Transformer and CNN encoding models.}
    \label{fig:benchmarks}
\end{figure*}

\section{Experiments on Transfer Tasks}

For comparison with prior USE models, English task transfer performance is evaluated on SentEval~\cite{conneau2018senteval}. For sentence classification transfer tasks, the output of the sentence encoders are provided to a task specific DNN. For the pairwise semantic similarity task, the similarity of sentence embeddings $u$ and $v$ is assessed using $-\arccos\Big(\frac{u v}{||u|| ~ ||v||}\Big)$ following \citet{yang2018}. As shown in table \ref{tab:trans-model-performance}, our multilingual models show competitive transfer performance comparing with state-of-the-art sentence embedding models.
USE\textsubscript{Trans} performs better than USE\textsubscript{CNN} in all tasks. Our new multilingual USE\textsubscript{Trans} even outperforms our best previously released English only model, USE\textsubscript{Trans} for English~\cite{use2018}, on some tasks.

\section{Resource Usage}

Figure (\ref{fig:benchmarks}) provides compute and memory usage benchmarks for our models.\footnote{
CPU benchmarks are run on Intel(R) Xeon(R) Platinum 8173M CPU @ 2.00GHz. GPU benchmarks were run on an NVidia v100. Memory footprint was measured on CPU.
}
Inference times on GPU are 2 to 3 times faster than CPU. Our CNN models have the smallest memory footprint and are the fastest on both CPU and GPU.
The memory requirements increase with sentence length, with the Transformer model increasing more than twice as fast as the CNN model.\footnote{Transformer models are ultimately governed by a time and space complexity of $O(n^2)$. The benchmarks show for shorter sequence lengths the time and space requirements are dominated by computations that scale linearly with length and have a larger constant factor than the quadratic terms.}
While this makes CNNs an attractive choice for efficiently encoding longer texts, this comes with a corresponding drop in accuracy on many retrieval and transfer tasks.

\section{Conclusion}

We present two multilingual models for embedding sentence-length text. Our models \emph{embed text from 16 languages into a shared semantic embedding space} and achieve performance on transfer tasks that approaches monolingual sentence embedding models. The models achieve good performance on semantic retrieval (SR), bitext retrieval (BR) and retrieval question answering (ReQA). They achieve performance on cross-lingual semantic retrieval (cl-SR) and cross-lingual retrieval question answering (cl-ReQA) that approaches monolingual SR and ReQA performance for many language pars. Our models are made freely available with additional documentation and tutorial colaboratory notebooks at: {\href{https://tfhub.dev/s?q=universal-sentence-encoder-multilingual}{https://tfhub.dev/s?q=universal-sentence-}
\href{https://tfhub.dev/s?q=universal-sentence-encoder-multilingual}{encoder-multilingual}}.

\section*{Acknowledgments}

We thank our teammates from Descartes and other Google groups for their feedback and suggestions.
Special thanks goes to Muthu Chidambaram for his early exploration of multilingual training, Taku Kudo for the SentencePiece model support, Chen Chen for the templates used to perform the transfer learning experiments and Mario Guajardo for an early version of the ReQA tutorial Colab.

\small
\bibliography{emnlp-ijcnlp-2019}

\begin{thebibliography}{25}
\expandafter\ifx\csname natexlab\endcsname\relax\def\natexlab#1{#1}\fi

\bibitem[{Abadi et~al.(2016)Abadi, Barham, Chen, Chen, Davis, Dean, Devin,
  Ghemawat, Irving, Isard, Kudlur, Levenberg, Monga, Moore, Murray, Steiner,
  Tucker, Vasudevan, Warden, Wicke, Yu, and Zheng}]{tf2016}
Mart\'{\i}n Abadi, Paul Barham, Jianmin Chen, Zhifeng Chen, Andy Davis, Jeffrey
  Dean, Matthieu Devin, Sanjay Ghemawat, Geoffrey Irving, Michael Isard,
  Manjunath Kudlur, Josh Levenberg, Rajat Monga, Sherry Moore, Derek~G. Murray,
  Benoit Steiner, Paul Tucker, Vijay Vasudevan, Pete Warden, Martin Wicke, Yuan
  Yu, and Xiaoqiang Zheng. 2016.
\newblock \href {http://dl.acm.org/citation.cfm?id=3026877.3026899}
  {Tensorflow: A system for large-scale machine learning}.
\newblock In \emph{Proceedings of USENIX OSDI'16}, OSDI'16, pages 265--283.

\bibitem[{Ba et~al.(2016)Ba, Kiros, and Hinton}]{ba2016}
Lei~Jimmy Ba, Ryan Kiros, and Geoffrey~E. Hinton. 2016.
\newblock \href {http://arxiv.org/abs/1607.06450} {Layer normalization}.
\newblock \emph{CoRR}, abs/1607.06450.

\bibitem[{Bowman et~al.(2015)Bowman, Angeli, Potts, and
  Manning}]{bowman-etal-2015-large}
Samuel~R. Bowman, Gabor Angeli, Christopher Potts, and Christopher~D. Manning.
  2015.
\newblock \href {https://doi.org/10.18653/v1/D15-1075} {A large annotated
  corpus for learning natural language inference}.
\newblock In \emph{Proceedings of the 2015 Conference on Empirical Methods in
  Natural Language Processing}, pages 632--642.

\bibitem[{Cakaloglu et~al.(2018)Cakaloglu, Szegedy, and Xu}]{cakaloglu2018}
Tolgahan Cakaloglu, Christian Szegedy, and Xiaowei Xu. 2018.
\newblock \href {http://arxiv.org/abs/1810.10176} {Text embeddings for
  retrieval from a large knowledge base}.
\newblock \emph{CoRR}, abs/1810.10176.

\bibitem[{Cer et~al.(2017)Cer, Diab, Agirre, Lopez-Gazpio, and
  Specia}]{cer-etal-2017-semeval}
Daniel Cer, Mona Diab, Eneko Agirre, I{\~n}igo Lopez-Gazpio, and Lucia Specia.
  2017.
\newblock \href {https://doi.org/10.18653/v1/S17-2001} {{S}em{E}val-2017 task
  1: Semantic textual similarity multilingual and crosslingual focused
  evaluation}.
\newblock In \emph{Proceedings of the 11th International Workshop on Semantic
  Evaluation ({S}em{E}val-2017)}, pages 1--14.

\bibitem[{Cer et~al.(2018)Cer, Yang, Kong, Hua, Limtiaco, St.~John, Constant,
  Guajardo-Cespedes, Yuan, Tar, Strope, and Kurzweil}]{use2018}
Daniel Cer, Yinfei Yang, Sheng-yi Kong, Nan Hua, Nicole Limtiaco, Rhomni
  St.~John, Noah Constant, Mario Guajardo-Cespedes, Steve Yuan, Chris Tar,
  Brian Strope, and Ray Kurzweil. 2018.
\newblock \href {https://www.aclweb.org/anthology/D18-2029} {Universal sentence
  encoder for {E}nglish}.
\newblock In \emph{Proceedings of the 2018 Conference on Empirical Methods in
  Natural Language Processing: System Demonstrations}, pages 169--174.

\bibitem[{Chen et~al.(2017)Chen, Fisch, Weston, and
  Bordes}]{chen-etal-2017-reading}
Danqi Chen, Adam Fisch, Jason Weston, and Antoine Bordes. 2017.
\newblock \href {https://doi.org/10.18653/v1/P17-1171} {Reading {W}ikipedia to
  answer open-domain questions}.
\newblock In \emph{Proceedings of the 55th Annual Meeting of the Association
  for Computational Linguistics (Volume 1: Long Papers)}, pages 1870--1879.

\bibitem[{Chidambaram et~al.(2018)Chidambaram, Yang, Cer, Yuan, Sung, Strope,
  and Kurzweil}]{mutty2018}
Muthuraman Chidambaram, Yinfei Yang, Daniel Cer, Steve Yuan, Yun{-}Hsuan Sung,
  Brian Strope, and Ray Kurzweil. 2018.
\newblock \href {http://arxiv.org/abs/1810.12836} {Learning cross-lingual
  sentence representations via a multi-task dual-encoder model}.
\newblock \emph{CoRR}, abs/1810.12836.

\bibitem[{Conneau and Kiela(2018)}]{conneau2018senteval}
Alexis Conneau and Douwe Kiela. 2018.
\newblock \href {https://www.aclweb.org/anthology/L18-1269} {{S}ent{E}val: An
  evaluation toolkit for universal sentence representations}.
\newblock In \emph{Proceedings of the Eleventh International Conference on
  Language Resources and Evaluation ({LREC}-2018)}.

\bibitem[{Conneau et~al.(2017)Conneau, Kiela, Schwenk, Barrault, and
  Bordes}]{conneau2017}
Alexis Conneau, Douwe Kiela, Holger Schwenk, Lo{\"\i}c Barrault, and Antoine
  Bordes. 2017.
\newblock \href {https://doi.org/10.18653/v1/D17-1070} {Supervised learning of
  universal sentence representations from natural language inference data}.
\newblock In \emph{Proceedings of the 2017 Conference on Empirical Methods in
  Natural Language Processing}, pages 670--680, Copenhagen, Denmark.
  Association for Computational Linguistics.

\bibitem[{Dolan et~al.(2004)Dolan, Quirk, and
  Brockett}]{dolan-etal-2004-unsupervised}
Bill Dolan, Chris Quirk, and Chris Brockett. 2004.
\newblock \href {https://www.aclweb.org/anthology/C04-1051} {Unsupervised
  construction of large paraphrase corpora: Exploiting massively parallel news
  sources}.
\newblock In \emph{{COLING} 2004: Proceedings of the 20th International
  Conference on Computational Linguistics}, pages 350--356.

\bibitem[{Gillick et~al.(2018)Gillick, Presta, and
  Tomar}]{Gillick2018EndtoEndRI}
Daniel Gillick, Alessandro Presta, and Gaurav~Singh Tomar. 2018.
\newblock \href {http://arxiv.org/abs/1811.08008} {End-to-end retrieval in
  continuous space}.
\newblock \emph{CoRR}, abs/1811.08008.

\bibitem[{Hoogeveen et~al.(2015)Hoogeveen, Verspoor, and Baldwin}]{quora}
Doris Hoogeveen, Karin~M. Verspoor, and Timothy Baldwin. 2015.
\newblock \href {https://doi.org/10.1145/2838931.2838934} {Cqadupstack: A
  benchmark data set for community question-answering research}.
\newblock In \emph{Proceedings of the 20th Australasian Document Computing
  Symposium}, ADCS '15, pages 3:1--3:8.

\bibitem[{Jones et~al.(2000)Jones, Walker, and Robertson}]{bm25}
K.~Sparck Jones, S.~Walker, and S.~E. Robertson. 2000.
\newblock \href {https://doi.org/10.1016/S0306-4573(00)00015-7} {A
  probabilistic model of information retrieval: Development and comparative
  experiments}.
\newblock \emph{Inf. Process. Manage.}, 36(6):779--808.

\bibitem[{Kim(2014)}]{kim-2014-convolutional}
Yoon Kim. 2014.
\newblock \href {https://doi.org/10.3115/v1/D14-1181} {Convolutional neural
  networks for sentence classification}.
\newblock In \emph{Proceedings of the 2014 Conference on Empirical Methods in
  Natural Language Processing ({EMNLP})}, pages 1746--1751.

\bibitem[{Kudo and Richardson(2018)}]{spm}
Taku Kudo and John Richardson. 2018.
\newblock \href {https://www.aclweb.org/anthology/D18-2012} {{S}entence{P}iece:
  A simple and language independent subword tokenizer and detokenizer for
  neural text processing}.
\newblock In \emph{Proceedings of the 2018 Conference on Empirical Methods in
  Natural Language Processing: System Demonstrations}, pages 66--71.

\bibitem[{Lei et~al.(2016)Lei, Joshi, Barzilay, Jaakkola, Tymoshenko,
  Moschitti, and M{\`a}rquez}]{askubuntu}
Tao Lei, Hrishikesh Joshi, Regina Barzilay, Tommi Jaakkola, Kateryna
  Tymoshenko, Alessandro Moschitti, and Llu{\'\i}s M{\`a}rquez. 2016.
\newblock \href {https://doi.org/10.18653/v1/N16-1153} {Semi-supervised
  question retrieval with gated convolutions}.
\newblock In \emph{Proceedings of the 2016 Conference of the North {A}merican
  Chapter of the Association for Computational Linguistics: Human Language
  Technologies}, pages 1279--1289.

\bibitem[{Logeswaran and Lee(2018)}]{quickthought2018}
Lajanugen Logeswaran and Honglak Lee. 2018.
\newblock \href {https://openreview.net/forum?id=rJvJXZb0W} {An efficient
  framework for learning sentence representations}.
\newblock In \emph{International Conference on Learning Representations
  (ICLR)}.

\bibitem[{Rajpurkar et~al.(2016)Rajpurkar, Zhang, Lopyrev, and Liang}]{squad}
Pranav Rajpurkar, Jian Zhang, Konstantin Lopyrev, and Percy Liang. 2016.
\newblock \href {https://doi.org/10.18653/v1/D16-1264} {{SQ}u{AD}: 100,000+
  questions for machine comprehension of text}.
\newblock In \emph{Proceedings of the 2016 Conference on Empirical Methods in
  Natural Language Processing}, pages 2383--2392.

\bibitem[{Uszkoreit et~al.(2010)Uszkoreit, Ponte, Popat, and
  Dubiner}]{jakob2010}
Jakob Uszkoreit, Jay Ponte, Ashok Popat, and Moshe Dubiner. 2010.
\newblock \href {https://www.aclweb.org/anthology/C10-1124} {Large scale
  parallel document mining for machine translation}.
\newblock In \emph{Proceedings of the 23rd International Conference on
  Computational Linguistics (Coling 2010)}, pages 1101--1109.

\bibitem[{Vaswani et~al.(2017)Vaswani, Shazeer, Parmar, Uszkoreit, Jones,
  Gomez, Kaiser, and Polosukhin}]{vaswani2017}
Ashish Vaswani, Noam Shazeer, Niki Parmar, Jakob Uszkoreit, Llion Jones,
  Aidan~N Gomez, \L~ukasz Kaiser, and Illia Polosukhin. 2017.
\newblock \href
  {http://papers.nips.cc/paper/7181-attention-is-all-you-need.pdf} {Attention
  is all you need}.
\newblock In \emph{Proceedings of NIPS}, pages 6000--6010.

\bibitem[{Williams et~al.(2018)Williams, Nangia, and
  Bowman}]{williams-etal-2018-broad}
Adina Williams, Nikita Nangia, and Samuel Bowman. 2018.
\newblock \href {https://doi.org/10.18653/v1/N18-1101} {A broad-coverage
  challenge corpus for sentence understanding through inference}.
\newblock In \emph{Proceedings of the 2018 Conference of the North {A}merican
  Chapter of the Association for Computational Linguistics: Human Language
  Technologies, Volume 1 (Long Papers)}, pages 1112--1122.

\bibitem[{Yang et~al.(2019)Yang, {\'{A}}brego, Yuan, Guo, Shen, Cer, Sung,
  Strope, and Kurzweil}]{Yang2019ImprovingMS}
Yinfei Yang, Gustavo~Hern{\'{a}}ndez {\'{A}}brego, Steve Yuan, Mandy Guo,
  Qinlan Shen, Daniel Cer, Yun{-}Hsuan Sung, Brian Strope, and Ray Kurzweil.
  2019.
\newblock \href {http://arxiv.org/abs/1902.08564} {Improving multilingual
  sentence embedding using bi-directional dual encoder with additive margin
  softmax}.
\newblock \emph{CoRR}, abs/1902.08564.

\bibitem[{Yang et~al.(2018)Yang, Yuan, Cer, Kong, Constant, Pilar, Ge, Sung,
  Strope, and Kurzweil}]{yang2018}
Yinfei Yang, Steve Yuan, Daniel Cer, Sheng-yi Kong, Noah Constant, Petr Pilar,
  Heming Ge, Yun-Hsuan Sung, Brian Strope, and Ray Kurzweil. 2018.
\newblock \href {https://www.aclweb.org/anthology/W18-3022} {Learning semantic
  textual similarity from conversations}.
\newblock In \emph{Proceedings of The Third Workshop on Representation Learning
  for {NLP}}, pages 164--174.

\bibitem[{Ziemski et~al.(2016)Ziemski, Junczys-Dowmunt, and
  Pouliquen}]{uncorpus}
Micha{\l} Ziemski, Marcin Junczys-Dowmunt, and Bruno Pouliquen. 2016.
\newblock \href {https://www.aclweb.org/anthology/L16-1561} {The united nations
  parallel corpus v1.0}.
\newblock In \emph{Proceedings of the Tenth International Conference on
  Language Resources and Evaluation ({LREC} 2016)}, pages 3530--3534.

\end{thebibliography}
\bibliographystyle{acl_natbib}

\appendix

\end{document}